\pdfoutput=1
\documentclass[10pt, a4paper]{article}
\usepackage[]{lrec2026} 
\usepackage{times}
\usepackage{latexsym}
\usepackage[T1]{fontenc}
\usepackage[utf8]{inputenc}
\usepackage{microtype}
\usepackage{enumitem}
\usepackage{booktabs}
\usepackage{multirow}
\usepackage{comment}
\usepackage{tcolorbox}
\usepackage{tikz}
\definecolor{mygreen}{HTML}{26A69A}

\usepackage{inconsolata}
\usepackage{graphicx}
\usepackage{subcaption}
 
\usepackage{tipa}
\newcommand{\umelb}{\textrm{\normalfont \textipa{@}}}
\newcommand{\uzur}{\textrm{\normalfont \textipa{Z}}}

\title{CommonMorph: Participatory Morphological Documentation Platform}

\name{Aso Mahmudi$^\umelb$, Sina Ahmadi$^\uzur$, Kemal Kurniawan$^\umelb$, 
\\ {\bf \large Rico Sennrich$^\uzur$, Eduard Hovy$^\umelb$, Ekaterina Vylomova$^\umelb$} }

\address{$^\umelb$The University of Melbourne,  $^\uzur$University of Zurich \\
         \{mahmudia, vylomovae\}@unimelb.edu.au}

\abstract{
Collecting and annotating morphological data present significant challenges, requiring linguistic expertise, methodological rigour, and substantial resources. These barriers are particularly acute for low-resource languages and varieties. To accelerate this process, we introduce \texttt{CommonMorph}, a comprehensive platform that streamlines morphological data collection development through a three-tiered approach: expert linguistic definition, contributor elicitation, and community validation. The platform minimises manual work by incorporating active learning, annotation suggestions, and tools to import and adapt materials from related languages. It accommodates diverse morphological systems, including fusional, agglutinative, and root-and-pattern morphologies. Its open-source design and UniMorph-compatible outputs ensure accessibility and interoperability with NLP tools.
Our platform is accessible at \url{https://common-morph.com}, offering a replicable model for preserving linguistic diversity through collaborative technology.
\\ \newline \Keywords{Morphology, UniMorph, Multilingual Resources, Low-Resource Languages} }

\begin{document}
\maketitleabstract

\section{Introduction}

With over 1,500 languages at risk of extinction by 2100 \citep{bromham2022global}, scaling and improving the efficiency of language documentation and data collection is essential to prevent further irreversible losses. During the ``International Decade of Indigenous Languages'' (2022--2032)~\citep{un_indigenous_2019}, tools that assist linguists and communities in collecting linguistic data for long-term documentation and preservation should be a priority. Morphological data collection and analysis, the systematic documentation of word structure and inflection patterns, remains one of the most labour-intensive steps of language documentation~\citep{ginn_findings_2023}, particularly for low-resource languages with complex inflectional systems. 

Finite-state transducers (FSTs), the most established rule-based approach to model morphologies~\citep{gorman2022finite}, are capable of generating and analysing wordforms licensed by a given grammar. However, the design and implementation of FSTs are labour-intensive, requiring extensive linguistic engineering expertise and a high degree of manual effort \citep{beemer-etal-2020-linguist}. Handling irregularities often requires ad hoc solutions, while maintaining quality across large-scale FSTs remains difficult. These factors make it challenging to scale FST-based approaches to the hundreds of under-documented languages urgently requiring attention. Moreover, FSTs do not generalise effectively to out-of-vocabulary (OOV) forms, which limits their applicability in many real-world documentation and processing scenarios.

\begin{figure}[t]
\centering
\includegraphics[width=1\linewidth]{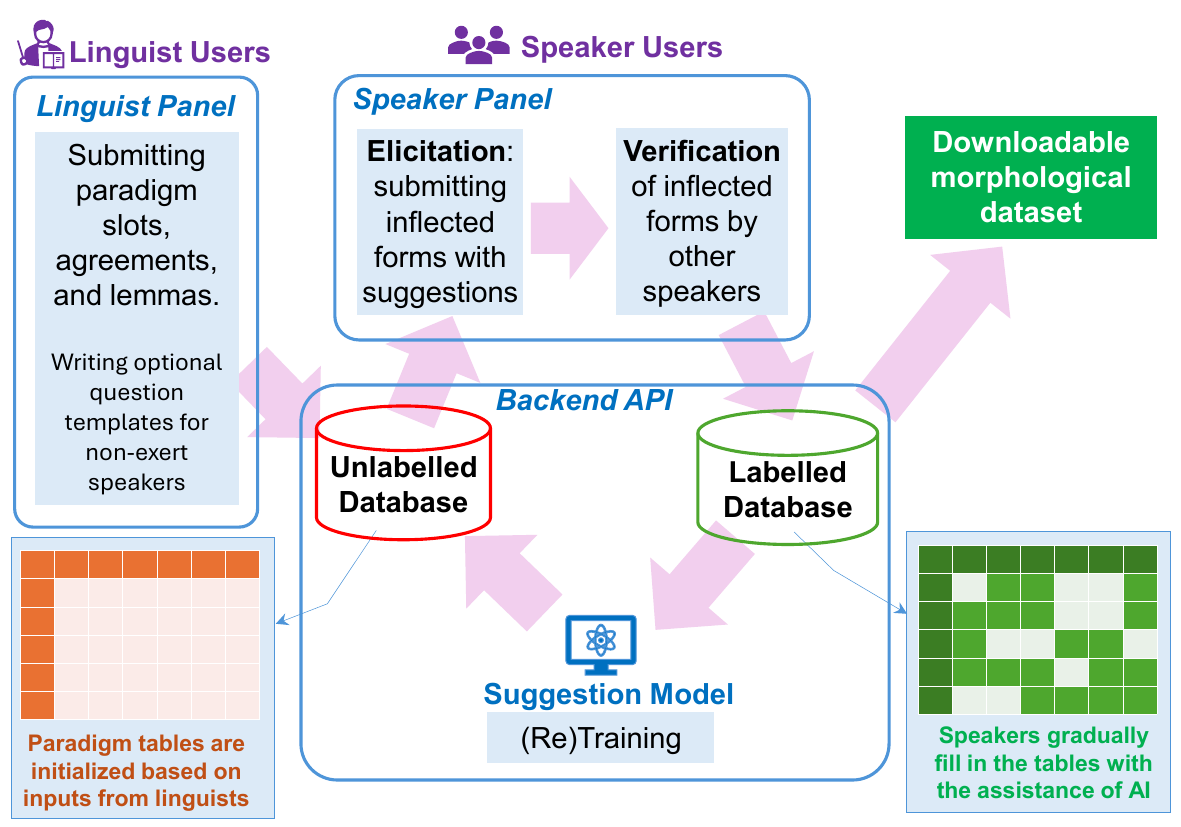}
\caption{The \texttt{CommonMorph} platform workflow facilitates elaboration of morphological structures by a linguist and provides an interoperable ecosystem for contributors to validate and enrich labelled databases.}
\label{fig:architecture}
\end{figure}

Modern data-driven approaches to morphological analysis and generation rely on large volumes of annotated morphological data—resources that are rarely available for low-resource or endangered languages. The UniMorph project \citep{batsuren-etal-2022-unimorph}, while the most comprehensive of its kind, provides datasets for only 169 languages, or less than 2.5\% of the world's languages. UniMorph has advanced computational morphology and enabled SIGMORPHON shared tasks on morphological segmentation \citep{batsuren-etal-2022-sigmorphon}, analysis~\citep{mccarthy-etal-2019-sigmorphon}, and generation~\citep{kann-schutze-2018-neural}, but faces key limitations:
(a)~the data preparation process \citep{kirov-etal-2018-unimorph} is time-consuming and not easily replicable across languages;
(b) systematic correction and maintenance require substantial resources; and
(c)~validation is limited, as there are no mechanisms for broad community feedback. For example, \citet{ahmadi-mahmudi-2023-revisiting} identified and corrected 25\% of the misleading forms in the Central Kurdish dataset.

A further limitation across existing approaches is the lack of systematic integration of related language varieties. Since most languages belong to a few hundred families \citep{campbell2018many}, related varieties often share morphological, phonological, and lexical patterns. Exploiting these similarities can reduce annotation effort and accelerate documentation, yet current tools rarely support such reuse.

This paper presents \texttt{CommonMorph}, a platform designed to streamline and scale the collection of inflectional morphological data across languages. By connecting linguists and speakers in a shared digital environment, \texttt{CommonMorph} supports collaborative annotation of wordforms with morphosyntactic features. While the system is applicable to all languages, it is particularly well-suited to low-resource and endangered ones, whose diverse and complex morphological systems require adaptable and scalable tools. Unlike traditional fieldwork practices that rely on isolated and offline contributions, the platform enables structured data entry, review, and community validation. It also enhances efficiency through imports from related languages and predictive suggestions. Developed iteratively through user feedback, \texttt{CommonMorph} is tailored to the workflows of both linguists and speaker communities.

To demonstrate its broad applicability, we tested \texttt{CommonMorph} on languages from diverse families, including Indo-European (Italic, Germanic, Iranic), Semitic, Bantu, and Turkic. The case studies show that the platform can handle various morphological systems, from fusional to agglutinative to root-and-pattern structures.

In short, \texttt{CommonMorph} offers the following core features:
\begin{enumerate}[noitemsep,nolistsep]
\item \textbf{Intuitive Interface}: Developed in collaboration with both linguists and speakers, the platform is intuitive and easy to use, with high user satisfaction reported in surveys (§~\ref{sec:results}).
\item \textbf{Linguist-led Workflow}: Linguists can define initial materials and create patterns to reduce the workload for speakers, improving the efficiency of the early annotation process (§~\ref{linguist-UI}). These materials can be adapted from existing resources of related languages in the system.
\item \textbf{Built-in Validation}: The platform's interactive evaluation of suggestions allows users to iteratively refine the data (§~\ref{speaker-UI}).
\item \textbf{Compatibility}: Outputs can be exported in standard formats such as UniMorph \citep{batsuren-etal-2022-unimorph}, enabling integration with existing NLP tools.
\item \textbf{Free and Open-Source}: \texttt{CommonMorph} is accessible to all and openly available for further development and adaptation.\footnote{Website: \url{https://common-morph.com}, the source code is available at \url{https://github.com/Aso-UniMelb/CommonMorph}}
\end{enumerate}

\section{Related Work} \label{sec:literature}

Recent work in computational linguistics has explored alternative methods for collecting morphological data, ranging from gamified participation to automated annotation. Gamification has been applied to morphological analysis, education, and community engagement, showing potential for improving data coverage and speaker involvement~\citep{eryiugit2023gamification,qiao2023understanding,sun-etal-2025-dia}. The rise of large language models (LLMs) has also enabled new forms of automatic linguistic annotation across a wide variety of tasks~\citep{tan-etal-2024-large}.

Despite these innovations, the development of high-quality morphological data and tools still depends heavily on manual work. \citet{beemer-etal-2020-linguist} estimate that developing a finite-state transducer (FST) analyser and generator comparable to a neural model requires an average of 40 person-hours, excluding infrastructure and design time. The effort varies substantially across languages: while regular inflectional systems can be implemented in under an hour, more complex ones demand extensive analysis of paradigms, lemma classes, and morphophonological alternations. These figures assume prior expertise, underscoring the difficulty of scaling traditional approaches.

To mitigate annotation costs, active learning methods have been explored for morphological segmentation and labelling, where algorithms iteratively select the most informative examples for human review \citep{baldridge-palmer-2009-well,palmer_semi-automated_2009}. Building on this idea, \citet{mahmudi-etal-2025-neural} propose techniques to help field linguists detect and prioritise data gaps for more efficient elicitation.

A major effort in large-scale morphological data collection is the UniMorph project. The first release by \citet{kirov-etal-2016-large} relied on manual mapping of Wiktionary grammatical descriptors to the UniMorph schema. In the second release report, \citet{kirov-etal-2018-unimorph} showed that verifying and correcting automatically parsed tables required at least as much effort as direct manual annotation, though annotators successfully produced initial datasets for 47 languages within days. The third phase \citep{mccarthy-etal-2020-unimorph} expanded coverage to 50 more languages with a team of nine annotators.

Data validation within UniMorph evolved from manual expert review to more systematic quality checks. Early annotations were adjudicated by linguists or speakers \citep{kirov-etal-2018-unimorph}, while later efforts integrated automated validation tools to flag errors such as rare characters, malformed cells, or inconsistencies between lemma and inflected form~\citep{mccarthy-etal-2020-unimorph}. These steps improved quality control but still relied heavily on post-hoc correction.

Despite these advances, to the best of our knowledge, no existing morphological data collection tool provides a unified platform that integrates structured workflows involving both linguists and speakers while supporting remote collaboration. Our \texttt{CommonMorph} platform fills this gap by combining user-friendly interfaces, active speaker participation, and features for guided elicitation. It supports both manual and semi-automated annotation and encourages the use of cross-linguistic data to reduce redundancy and improve scalability.

\section{Design Principles}
The \texttt{CommonMorph} platform is designed to address the core research question: \textit{How can morphological data collection be streamlined and scaled for low-resource and endangered languages through collaboration between linguists and speakers?} To achieve this, the platform adheres to principles that balance the needs of its primary contributors, linguists and speakers \citep{flavelle-lachler-2023-strengthening}, while accommodating diverse morphological systems.

Linguists play a central role in defining the scope and quality of morphological resources. 
The platform enables them to define paradigms, morphosyntactic features, and lexical entries across diverse language varieties with fine-grained control. They can follow established annotation standards while also adding community-accepted aliases for grammatical terms. Materials from related languages can be imported and adapted, allowing cross-linguistic regularities to reduce redundant work. Linguists can also revise and expand data as new insights emerge, ensuring that resources remain linguistically accurate and up to date. These capabilities ensure that expert knowledge anchors the data collection process and maintains its reliability.

Speakers are the primary custodians of endangered and low-resource languages. To empower non-experts to participate, the platform offers an intuitive interface that enables speakers to annotate and validate forms without requiring specialist training. Its flexible workflows adapt to varying levels of grammatical knowledge within the community, while community feedback loops capture corrections, variations, and alternative forms. Active learning-driven suggestions for inflected forms, based on patterns in the data, help reduce manual effort. This principle recognises speakers not merely as informants but as active collaborators, fostering inclusivity and shared ownership of the resource.

\section{System Design}
The \texttt{CommonMorph} platform, as depicted in Figure~\ref{fig:architecture}, begins with linguists defining initial linguistic materials. Based on this setup, the system generates suggested inflected forms, which speakers review by correcting or confirming them. As verified data accumulates, the system improves its suggestions through active learning, reducing manual effort over time. All data remains editable and is exportable in standard formats for integration with other linguistic and NLP tools. In the following subsections, we provide a detailed description of our approach.

\subsection{Terminology}

Before outlining the system design, we clarify the terminology used throughout the platform:
\begin{itemize}[noitemsep,nolistsep]
\item A \textbf{lemma} (e.g., walk) is the base dictionary form representing a set of inflected wordforms (e.g., walks, walking).
\item An \textbf{inflection class} groups lemmas that share the same inflectional pattern. For instance, most English regular verbs follow a four-form pattern (walk-s, walk-ing, walk-ed, walk-ed (past participle)).
\item A \textbf{paradigm} is the complete set of inflected forms derived from a single lemma, typically organised according to morphosyntactic features such as tense, number, or person.
\item A \textbf{gloss} is a short explanation or translation that conveys the meaning or function of a form.
\item A \textbf{morphosyntactic feature set} is a bundle of grammatical properties such as tense, person, or number. For consistency, \texttt{CommonMorph} adopts the UniMorph schema~\citep{sylak2016composition} for glossing these features.
\item \textbf{Morphophonological rules} describe spelling changes that occur when morphemes combine in particular contexts (e.g., the plural of bus becomes buses).
\end{itemize}

\subsection{Required Linguistic Materials} \label{data}

We assume the availability of some linguistic materials, either gathered during early fieldwork stages or sourced from previous descriptive resources. Note that we anticipate these materials may contain errors, and linguists should have the ability to correct or expand them at any time. These materials should include:
\begin{enumerate}[noitemsep,nolistsep]
 \item \textbf{Basic wordlist or vocabulary} \citep[Sec. 3.1.3]{bowern_linguistic_2015} covering the inflection classes of the language, and
 \item \textbf{Morphosyntactic feature combinations} for each inflection class, typically organised in paradigms or conjugation tables. These combinations describe how lexical items vary according to grammatical categories such as person, number, tense, aspect, mood, or case. Figure~\ref{fig:conjugation} provides an example of such a resource.
\end{enumerate}
In practice, documentation often begins with incomplete information that must be iteratively refined. This initial data may be sourced from prior descriptive studies or inferred from related language varieties. While linguistic resources exist for many languages, including endangered ones, their structure and quality vary widely. Some wordlists and lexicons are highly detailed, while others lack even basic part-of-speech tags. Similarly, morphosyntactic features can appear as detailed grammatical descriptions or as simple tabular paradigms. To manage this variation, linguists are guided to enter and standardise the materials into our platform in a consistent and unified format.

\subsection{User Roles}

In traditional linguistic fieldwork, the linguist typically guides the elicitation process based on their expertise. In a digital environment, however, managing this interaction and ensuring that contributors enter accurate data can be challenging. We designed the \texttt{CommonMorph} user interface to accommodate users with varying levels of familiarity with linguistic concepts.

\texttt{CommonMorph} considers two roles of users:
\begin{itemize}
 \item \textbf{Linguists} (or language enthusiasts with morphological expertise) initiate a project by providing lemmas, paradigm structures (§\ref{data}), and optionally inflectional patterns and annotation priorities. They oversee data quality, guide the system's predictions, and ensure consistency.
 \item \textbf{Speakers} (informants or consultants without formal linguistic training) contribute by supplying inflected forms with hints from system-generated suggestions and verifying entries submitted by other speakers, enabling community-driven data validation.
\end{itemize}

Role-specific interfaces ensure that expert guidance and community participation are both supported. In the following subsections, we describe the user interface tailored for each user role.

\begin{figure}
    \centering
    \begin{subfigure}[b]{0.22\textwidth}        
        \setlength{\fboxrule}{0.1pt}
        \fcolorbox{gray}{white}{\includegraphics[width=\linewidth]{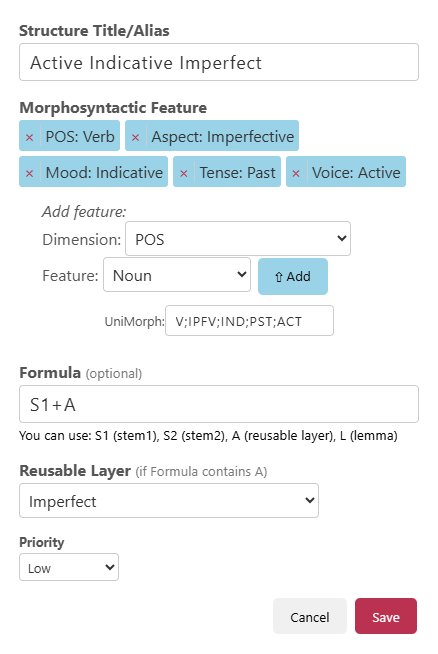}}
        \caption{Paradigm Structures}
        \label{fig:image1}
    \end{subfigure}
    \hfill
    \begin{subfigure}[b]{0.22\textwidth}
        \setlength{\fboxrule}{0.1pt}
        \fcolorbox{gray}{white}{\includegraphics[width=\linewidth]{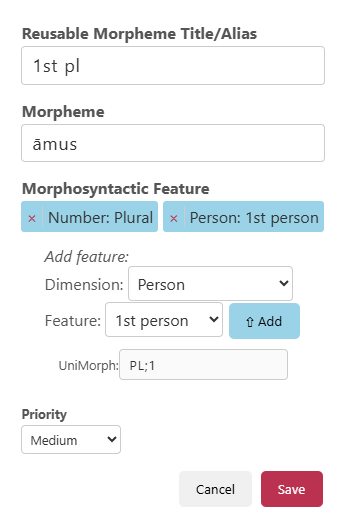}}
        \caption{Reusable Layers}
        \label{fig:image2}
    \end{subfigure}
    \hfill
    \begin{subfigure}[b]{0.22\textwidth}
        \setlength{\fboxrule}{0.1pt}
        \fcolorbox{gray}{white}{\includegraphics[width=\linewidth]{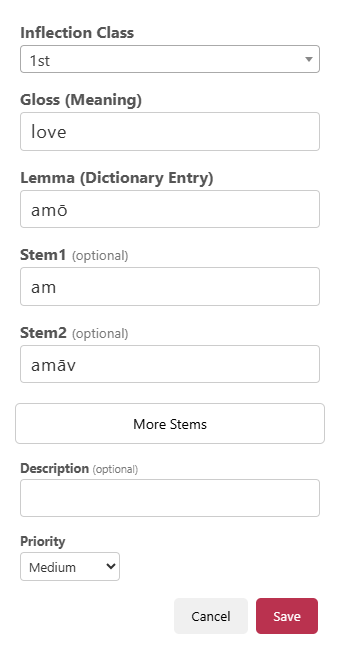}}
        \caption{Lexicon}
        \label{fig:image3}
    \end{subfigure}
    \hfill
    \begin{subfigure}[b]{0.22\textwidth}
        \setlength{\fboxrule}{0.1pt}
        \fcolorbox{gray}{white}{\includegraphics[width=\linewidth]{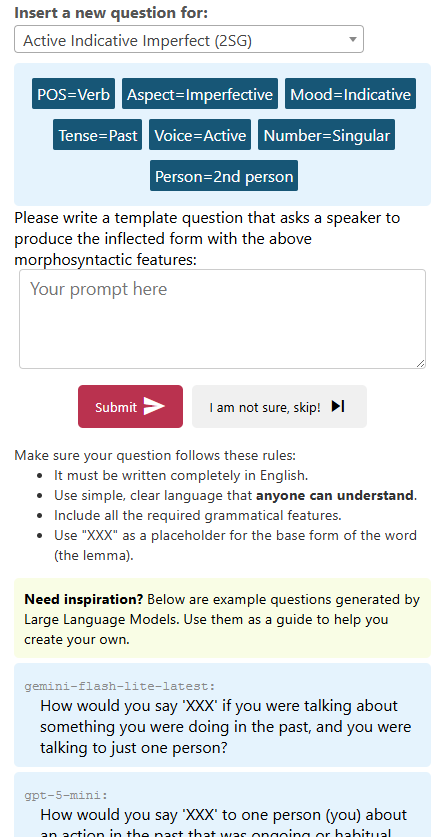}}
        \caption{Question Design}
        \label{fig:image4}
    \end{subfigure}
    \caption{Linguists define (a) paradigm structures, (b) additional structural layers such as agreement, and (c) lexical entries. These components allow the generation of morphological forms that can be validated during elicitation. Linguists also create elicitation prompts (d), taking inspiration from examples produced by the system.}
    \label{fig:linguistPanels}
\end{figure}

\subsection{Linguist Interface}  \label{linguist-UI}
The linguist interface is structured in a modular fashion to minimise redundancy and to exploit morphological regularities. It provides a workflow for defining paradigms, entering lexical items, and specifying rules, while allowing flexibility for irregular or complex systems. Figure~\ref{fig:linguistPanels} shows the interface during data submission. Core and optional modules include:

\begin{enumerate}[noitemsep]

 \item \textbf{Paradigm Structures} (Figure~\ref{fig:image1}): This core module defines the morphosyntactic structures that make up inflectional paradigms, such as tense, aspect, and mood. Each structure is tied to a specific inflection class. Optional inflectional patterns with placeholders allow the system to generate candidate forms before the machine learning models have sufficient data. 
 
 \item \textbf{Lexicon} (Figure~\ref{fig:image2}): Here, lemmas are entered with their assigned inflection class, stems, and glosses. Glosses ensure clarity for both contributors during elicitation and researchers during analysis.

\item \textbf{Reusable Layers} (Figure~\ref{fig:image3}): To avoid redundancy, this module allows defining reusable sets of features (e.g., agreement affixes) linked to paradigm structures. Simpler inflectional systems, such as English verbs, may not require this module.

 \item \textbf{Morphophonological rules (Optional)}: This module encodes orthographic adjustments (e.g., Turkish vowel harmony) as replacement patterns, implemented as regular expressions to reduce the need for manual corrections.
 
 \item \textbf{Question Design (Optional)} (Figure~\ref{fig:image4}): This module enables linguists to design elicitation prompts for non-expert speakers. Prompts are written in a shared meta-language (e.g., English or Spanish) familiar to both linguist and speaker, and structured as reusable templates with placeholders for lemmas. For example, the prompt \textit{``How do you tell more than one person to [LEMMA] right now?''} is used to generate a form with the following morphosyntactic features: POS=Verb, Mood=Imperative, Tense=Present, Number=Plural, and Person=2nd. Prompts can be created manually, adapted from other projects, or generated using LLMs (see Appendix~\ref{appendix_prompt_question} for the exact prompt template used for generation). This feature is especially valuable when speakers are unfamiliar with grammatical terminology.
\end{enumerate}

\begin{figure}
\centering
\setlength{\fboxrule}{0.1pt}
\fcolorbox{gray}{white}{\includegraphics[width=0.95\linewidth]{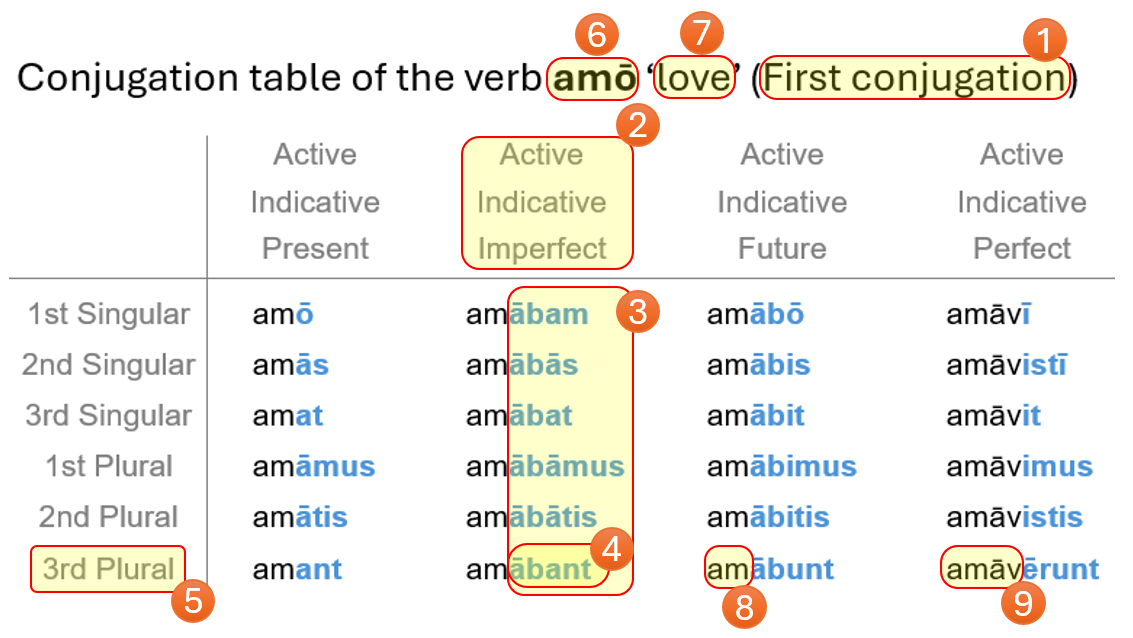}}
\caption{Example of a Latin verb conjugation table annotated with CommonMorph terminology:
(1) An \textit{inflection class} and (2) a \textit{paradigm structure} within that inflection class;
(3) a \textit{reusable layer} contains several (4) \textit{reusable morphemes} and their (5) morphosyntactic features;
(6) a \textit{lemma} and its (7) \textit{gloss} and (8–9) different \textit{stems}.}
\label{fig:conjugation}
\end{figure}

Figure~\ref{fig:conjugation} illustrates an example of the information linguists can add through the interface.
They can also assign priority levels to lemmas and structures, directing speakers to contribute the most informative data first.

To support scalability, cross-linguistic reuse, and community-driven resource sharing, the platform features import and export functionalities. Linguistic materials, including paradigm structures and lexicons, can be imported and exported in TSV format. This enables researchers to host and share resources openly and to adapt existing datasets from related languages with minimal effort.

\begin{figure}
    \centering
    \begin{subfigure}[b]{0.22\textwidth}
        \setlength{\fboxrule}{0.1pt}
        \fcolorbox{gray}{white}{\includegraphics[width=\linewidth]{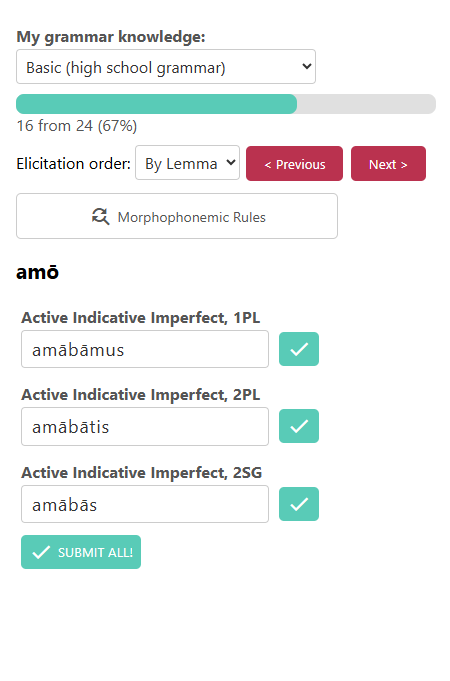}}
        \caption{Eliciting expert speakers}
    \end{subfigure}
    \hfill
    \begin{subfigure}[b]{0.22\textwidth}
        \setlength{\fboxrule}{0.1pt}
        \fcolorbox{gray}{white}{\includegraphics[width=\linewidth]{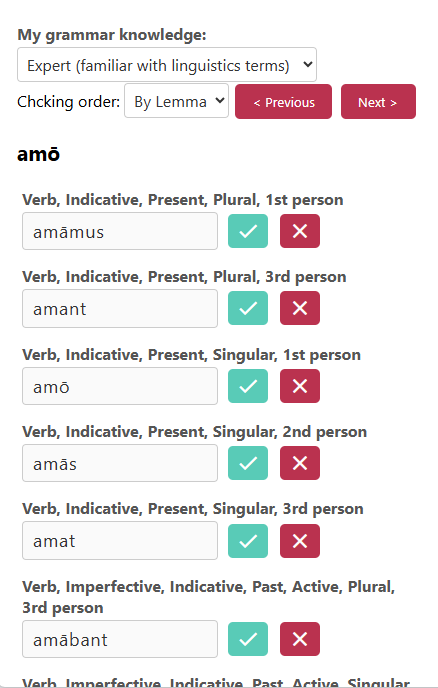}}
        \caption{Checking by expert speakers}
    \end{subfigure}
    \hfill
    \begin{subfigure}[b]{0.22\textwidth}
        \setlength{\fboxrule}{0.1pt}
        \fcolorbox{gray}{white}{\includegraphics[width=\linewidth]{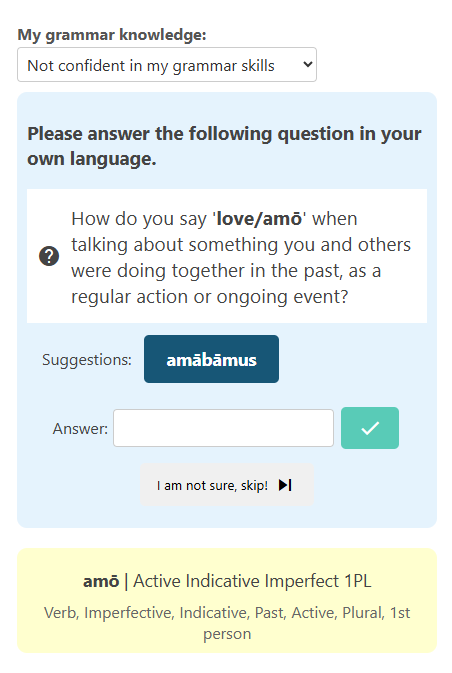}}
        \caption{Eliciting non-expert speakers}
    \end{subfigure}
    \hfill
    \begin{subfigure}[b]{0.22\textwidth}
        \setlength{\fboxrule}{0.1pt}
        \fcolorbox{gray}{white}{\includegraphics[width=\linewidth]{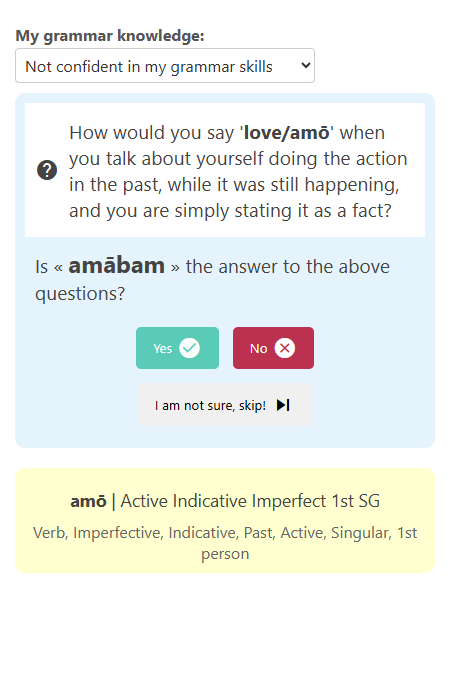}}
        \caption{Checking by non-expert speakers}
    \end{subfigure}
    \caption{Screenshots from the speaker interface.}
    \label{fig:speakerPanels}
\end{figure}

\subsection{Speaker Interface} \label{speaker-UI}
Once the linguist has provided the initial materials, the speaker panel becomes available for elicitation, as illustrated in Figure~\ref{fig:speakerPanels}. The panel supports two main activities: data entry (elicitation) and verification (review), with the interface adapting to the speaker's level of grammatical knowledge.

The system combines the defined structures with the lemmas to automatically generate paradigm tables that cover the full range of potential inflected forms. Since the number of possible forms can be very large, the platform presents speakers with only a subset, prioritised according to the linguist's annotation settings and the informativeness criteria outlined in Section~\ref{inflection_model}.

To provide transparency regarding how candidate morphological entries are computed, the interface explicitly informs the user about the source of each suggestion. Suggestions are visually tagged to indicate whether they were generated by the linguist's predefined rules, the internal neural inflection model, or an LLM.

Furthermore, because language documentation often reveals dialectal variation or simple human error, the review interface includes a mechanism for handling disagreements between speakers. If multiple speakers propose or validate conflicting forms for the same morphosyntactic slot, the platform flags the entry. These flagged disagreements are then either resolved through a majority consensus among designated expert speakers or escalated to the linguist for final adjudication.

The interface adapts to different levels of expertise: (1) Expert speakers see larger sets of forms, such as full or partial paradigm tables, which they can complete or verify in bulk. (2) Non-expert speakers see one form at a time with a simplified prompt that helps them provide or confirm the correct wordform without needing specialist terminology.

This adaptive design allows contributions from both experts and non-experts, combining detailed linguistic knowledge with broader community input.

\subsection{Inflection Model in Active Learning} \label{inflection_model}
The elicitation process follows an active learning framework \citep{mahmudi-etal-2025-neural}, progressing in iterative cycles to minimise annotation cost. \texttt{CommonMorph} utilises a dynamic, multi-source prediction pipeline that transitions from rule-based heuristics to data-driven models as the dataset grows.

\textbf{Cold Start:} Lacking initial training data, the system relies exclusively on linguist-supplied inflection patterns and morphophonological rules to generate early suggestions.

\textbf{Active Learning:} Validated forms periodically train a language-specific neural inflection model. This model drives elicitation by measuring predictive uncertainty to prioritise the ``most informative'' unannotated forms, while simultaneously generating increasingly accurate suggestions for speakers to review. 

\textbf{Integrating LLMs:} To bridge the low-data phase where custom neural models struggle, we leverage lower-latency LLMs (e.g., \texttt{Gemini-2.5-Flash}, \texttt{gpt-5-mini}) via API. The system dynamically constructs few-shot prompts using previously validated forms with matching morphosyntactic features as in-context examples (Appendix \ref{appendix_prompt_suggestion}), enabling accurate predictions without explicit fine-tuning.

\textbf{Ensemble Presentation:} The system aggregates suggestions from all three sources (rules, neural model, LLM). Converging predictions are presented with high confidence, while disagreements yield a multiple-choice list where each option is visually tagged by its generative source (Section~\ref{speaker-UI}). A comparative analysis of these models is provided in Section~\ref{sec:suggestion-models}.

\subsection{Implementation}
We implemented \texttt{CommonMorph} as a web application to make it freely accessible online, allowing researchers and practitioners to use it without installation barriers or platform restrictions. Since speakers of some endangered languages may not have stable internet access, the system includes a feature that allows linguists to print empty tables after submitting the initial data. These tables can then be filled out offline in remote areas and later integrated back into the system.

The backend and frontend are implemented as a .NET web application. All data is stored in a PostgreSQL database using relational tables to minimise redundancy. The application can be deployed with Docker and runs on both Linux and Windows servers. The active learning component is developed using FastAPI and PyTorch.

To implement the neural inflection model for active learning, we designed a simple and fast network that can be trained or fine-tuned on minimal server specifications. The model is a 2-layer LSTM with hidden and embedding dimensions of 128. All input sequences are padded or truncated to a fixed length of 20 character tokens, which enables efficient batch processing and stable training. The network is trained on a CPU for 15 epochs, taking approximately one minute, making it suitable for rapid experimentation in low-resource environments.

All development libraries and code used are free and open-source. For simplicity, we do not train or fine-tune LLMs locally; access to LLMs is provided via API.

\vspace{-3pt}
\section{Case studies}
To demonstrate the platform's ability to support languages with diverse morphological systems, we conducted a series of case studies covering a wide range of language families. The selected languages include Spanish and Latin (Italic), English and German (Germanic), Hawrami Kurdish, Central Kurdish, and Farsi (Iranic), Arabic (Semitic), Swahili (Bantu), and Turkish (Turkic).

Spanish, German, English, Farsi, and Kurdish primarily exhibit fusional suffixation with stem alternations. Hawrami additionally involves reduplication. Turkish employs agglutinative suffixation with vowel harmony, while Arabic features a templatic, non-concatenative morphology. 

The platform accommodates these typological differences through flexible paradigm definitions and stem management. In root-and-pattern systems, each lemma may specify multiple stems, allowing individual consonants of the root to be mapped to appropriate vocalic templates. Morphophonological rules further enable linguists to encode complex rewrite operations, such as the vowel harmony constraints observed in Turkish. By evaluating \texttt{CommonMorph} across these morphologically distinct systems, we assess its general applicability to typologically varied languages.

Furthermore, to assess the platform’s ability to support data collection using resources from related language varieties, we conducted experiments with several varieties from the Middle East. These experiments show how the platform can leverage linguistic similarity to assist under-resourced varieties. Data collection for each variety involved one linguist and one speaker. Initially, the standard varieties of Central Kurdish and Farsi were added to the platform. Afterwards, additional contributors added their own varieties, using the existing data as a reference when relevant.

Hawrami, an endangered language,\footnote{\url{https://unesdoc.unesco.org/ark:/48223/pf0000187026}} now has comprehensive digital data on the verbal morphology of its two dialects for the first time. We hope that these data will be used in future NLP applications to support the preservation of the language.

\begin{table*}
\centering \small
\setlength{\tabcolsep}{4pt}
\begin{tabular}{l| >{\raggedleft\arraybackslash}p{1.3cm}| >{\raggedleft\arraybackslash}p{1.3cm} | >{\raggedleft\arraybackslash}p{1.3cm}| >{\raggedleft\arraybackslash}p{1.5cm}| >{\raggedleft\arraybackslash}p{1.3cm}| >{\raggedleft\arraybackslash}p{1.5cm}}
\hline
\textbf{Suggestion Model}         &  Hawrami  {\footnotesize (Tewêle)} & Kurdish  {\footnotesize (Sine)} &  Farsi  {\footnotesize (Tehran)} & Turkish  {\footnotesize (Standard)} & Spanish  {\footnotesize (Spain)} &  Arabic  {\footnotesize (Standard)}\\ \hline
Linguist patterns  & 12.33 & 1.39 & 0.86 & 31.74 & 0.82 & 6.62 \\  
Linguist patterns + morphophonolgy  & \textbf{5.78} & \textbf{0.74} & \textbf{0.85} & 2.97 & 0.82 & \textbf{2.24} \\  \hline
NN trained on 100 samples          & 48.41 & 54.23 & 37.25 & 23.31 & 20.98 & 41.93 \\
NN trained on 500 samples          & 15.13 & 29.12 & 23.63 & 11.83 & 7.18 & 8.45 \\
NN trained on 1000 samples         & 7.88  & 27.37 & 54.02 & 3.33 & 5.91 & 6.40 \\
NN trained on 2000 samples         & 6.46  & 14.36 & 24.22 & 4.68 & 1.14  & 2.72 \\ \hline
\texttt{gemini-2.5-flash} (1-shot) & 17.00 & 14.86 & 3.71 & 2.36 & 0.50 & 7.25 \\
\texttt{gemini-2.5-flash} (2-shot) & 14.45 & 13.43 & 1.86 & 1.5 & 0.50 & 4.59 \\
\texttt{gemini-2.5-flash} (3-shot) & 13.03 &  6.86 & 2.32 & 2.15 & 0.50 & 5.07  \\
\texttt{gpt-5-mini} (1-shot)       & 19.26 & 15.14 & 4.64 & 0.43 & \textbf{0.00} & 10.14  \\
\texttt{gpt-5-mini} (2-shot)       & 11.61 & 14.57 & 4.64 & \textbf{0.21} & 0.25 & 7.97 \\ 
\texttt{gpt-5-mini} (3-shot)       & 12.18 &  7.43 & 3.25 & \textbf{0.21} & \textbf{0.00} & 4.83  \\ \hline
\end{tabular}
\caption{Character Error Rate (CER, in \%) of the inflection model in generating suggestions for different languages, which reflects the amount of correction required by speakers.}
\label{tab:suggestion-results}
\end{table*}

\section{Evaluation and Results} \label{sec:results}

It is difficult to establish a baseline for evaluating the efficiency of our platform because no existing system offers comparable functionalities. As discussed in Section~\ref{sec:literature}, some previous studies on morphological data collection have provided rough estimates of manual effort. However, we consider such comparisons inappropriate as our platform integrates a wider range of components and involves users with less specialised expertise.

To assess the effectiveness of \texttt{CommonMorph}, we conduct a multi-faceted evaluation focused on system performance and user experience. Our evaluation aligns with prior work in human-in-the-loop language documentation, where the focus is on reducing annotation effort while maintaining quality~\citep{palmer-etal-2009-evaluating, baldridge-palmer-2009-well}.

\subsection{Annotation Suggestion Accuracy} \label{sec:suggestion-models}
We employed an active learning strategy to improve the quality of annotation suggestions over time. The accuracy of model-generated suggestions was evaluated by comparing them to user-validated annotations using the Character Error Rate (CER) metric. Lower CER values indicate fewer corrections needed by annotators, and thus, higher model usefulness. 

Table~\ref{tab:suggestion-results} presents the Character Error Rate (CER) results for 3,000 randomly selected samples from several language varieties. These include Hawrami (Tewêle dialect), Central Kurdish (Sine dialect), Farsi (non-standard Tehrani dialect), Standard Turkish, Standard Spanish, and fully diacritized Modern Standard Arabic. For the first two datasets, no prior morphological data were available; they are published in \texttt{CommonMorph} for the first time.

We compare multiple suggestion model configurations, including linguist-provided patterns (with and without morphophonological rules), neural networks trained on varying amounts of data, and few-shot prompting with two LLMs. The results demonstrate that linguist-defined morphological patterns substantially enhance model performance for zero-resource varieties, underscoring the importance of expert linguistic knowledge in low-resource settings. Incorporating morphophonological rules further enhances suggestion accuracy for most varieties, particularly for Turkish, where vowel harmony plays a central role. However, providing detailed linguistic rules can be time-consuming, and in some cases, linguists may have limited knowledge of the target variety. In scenarios involving related languages where sufficient prior knowledge exists, the linguist user can instead focus on identifying potential points of deviation, reducing manual effort.

For neural network models, CER consistently decreases with additional training data, validating the effectiveness of active learning. Few-shot prompting with LLMs—particularly with two- or three-shot inputs—achieves competitive results for well-resourced standard languages. However, in the case of Arabic, model performance was lower due to the use of fully diacritised data, while most Arabic training data available to LLMs is undiacritised, leading to less accurate suggestions.

\subsection{User Survey}
To evaluate the usability and effectiveness of \texttt{CommonMorph}, we conducted a user survey involving both linguists and speakers. Participants reported their learning time (in minutes), contribution time (in hours), and rated the following aspects on a 5-point Likert scale: \textbf{ease of use} for clarity of the interface and task design, \textbf{efficiency} for users' perception of how to quickly and effectively complete tasks, \textbf{pedagogical value} for if the platform helps speakers and learners better understand the language structure, and \textbf{satisfaction} for overall user experience and willingness to reuse the platform. 

As shown in Table~\ref{tab:survey}, the platform received consistently high ratings across all categories, with average satisfaction and ease-of-use scores exceeding 4.7/5. Speakers reported shorter learning curves (8.7 minutes) compared to linguists (17.3 minutes), yet contributed substantially more time to the documentation effort. Both user groups gave strong pedagogical ratings, suggesting that the platform effectively supports language learning and awareness. The positive feedback indicates it helps bridge the expertise gap in collaborative documentation.

\begin{table}
\centering   \small
\setlength{\tabcolsep}{4pt}
\begin{tabular}{l|cc|cc}
\hline
~ & \multicolumn{2}{c}{Linguists} & \multicolumn{2}{c}{Speakers} \\
\textbf{Metric}        & Mean & SD & Mean & SD \\ \hline
User Learning Time     & 17.3 min & 2.5 & 8.7 min & 3.2 \\
Contributing Time      & 1.0 hr & 2.5 & 4.1 hr & 9.2 \\
Pedagogical Score      & 3.7/5  & 1.5 & 4.7/5  & 0.5 \\
Efficiency Score       & 4.3/5  & 1.1 & 4.7/5  & 0.5 \\
Satisfaction Score     & 4.7/5  & 0.5 & 4.7/5  & 0.5 \\
Ease of Use Score      & 4.7/5  & 0.5 & 5.0/5  & 0 \\ \hline
\end{tabular}
\caption{User survey results for the \texttt{CommonMorph} platform. The table reports average user learning and contribution times, as well as mean scores (1 = low, 5 = high) for pedagogical value, efficiency, satisfaction, and ease of use. SD indicates standard deviation.}
\label{tab:survey}
\end{table}

\section{Conclusion and Future Work}
This study presents \texttt{CommonMorph}, a collaborative platform designed to document and expand morphological resources. Our evaluation demonstrates that the platform effectively supports collaborative data collection by integrating rule-based linguistic input with machine learning suggestions derived from active learning. Both linguists and community speakers found the interface intuitive and easy to use, highlighting its potential for participatory language documentation.

A key outcome of this work is the demonstrated benefit of combining structured linguistic knowledge with the pattern-recognition capabilities of neural models. Initial morphological rules contributed by linguists substantially improved the quality of subsequent system-generated suggestions, reducing manual annotation effort while preserving data accuracy. This hybrid workflow proved especially valuable during the initial stages of data collection and enabled high-quality input from users with varying levels of expertise.

As the amount of collected data increases, the platform has the potential to offer a universal morphological analysis system. Additionally, because it captures glossing and morphosyntactic tags, it would assist linguists in generating interlinear glossed text (IGT) data.

Future work will focus on expanding both the scope and automation of the platform. One promising direction is the automatic extraction of morphological data and rule patterns from descriptive linguistic sources such as grammars and dictionaries~\citep{virk-etal-2020-dream}, allowing community reviewers to verify and refine the output. Extending the framework beyond morphology to other linguistic domains, including phonology, syntax, and semantics, would further enhance its usefulness for language documentation and computational modelling. Finally, because many endangered languages lack standardised orthographies, incorporating speech-based data entry, through manual verification or automatic speech recognition, will further enhance accessibility and inclusivity in language resource development.


\section*{Ethical Considerations}
The \texttt{CommonMorph} platform does not publish any personal or identifiable information from its users. All linguistic data contributed through the platform is intended for public release via open-access repositories. The study design and participant involvement have been reviewed and approved by the University of Melbourne Human Research Ethics Committee (Reference No. 2025-32066-64105-3). Our analyses are based on licensed data, which are freely available for academic use.

\section*{Acknowledgements}
We extend our deepest appreciation to the contributors of our case studies, especially Ako Marani, Mohammadreza Yadegari, Erfan Karami, and Morteza Naserzadeh for their invaluable time, dedication, and linguistic expertise. We are also deeply grateful to Borja Herce, John Mansfield, and Demian Aaron Inostroza Améstica for their helpful feedback and suggestions during the preparation of \texttt{CommonMorph}. Furthermore, we would like to thank the anonymous reviewers for their constructive feedback, which significantly improved the final version of this paper.

Sina Ahmadi gratefully thanks the support of the UZH Grant (reference number 269093). Ekaterina Vylomova expresses gratitude to the University of Melbourne for providing support for this research (ECR grant 2024ECRG140).

\section{Bibliographical References} \label{sec:reference}
\bibliographystyle{lrec2026-natbib}
\bibliography{ref}

\appendix
\counterwithin{figure}{section}
\counterwithin{table}{section}

\section{Prompt for Suggestion Generation} \label{appendix_prompt_suggestion}

The following prompt is employed to obtain a suggestion when two data samples from an identical morphosyntactic feature set have been submitted and verified by speakers. In languages like Kurdish, where verb lemmas often have two distinct stems, linguists can provide these stems in advance. By retrieving them from the database during prompting, the system can construct more detailed prompts, increasing the likelihood of generating accurate suggestions.

\begin{tcolorbox}[colback=lightgray!10, colframe=gray!50, title= A. Prompt for generating inflected forms suggestions, fonttitle=\small,fontupper=\small]
\texttt{In the language YYY, what is the correct inflected form of the lemma <LEMMA>, given that its stem1 is "<STEM1>" and stem2 is "<STEM2>" for a specific grammatical feature set? As reference examples, under the same grammatical features:
- The lemma <SAMPLE.1.LEMMA> (with stem1: "<SAMPLE.1.STEM.1>", stem2: "<SAMPLE.1.STEM.2>") yields the form <SAMPLE.1.FORM>.
- The lemma <SAMPLE.2.LEMMA> (with stem1: "<SAMPLE.2.STEM.1>", stem2: "<SAMPLE.2.STEM.2>") yields the form <SAMPLE.2.FORM>.
Think about this quietly and just give me one final inflected word.}
\end{tcolorbox}

For example, its placeholders will be replaced with the following words:
\begin{table}[h]
\centering
\resizebox{0.8\linewidth}{!}{
\begin{tabular}{ll}
\toprule
\textbf{Placeholder} & \textbf{Replaces with} \\ \hline
<LEMMA> & hênan \\
<STEM1> & hêna \\
<STEM2> & hên \\
<SAMPLE.1.LEMMA> & girtin \\
<SAMPLE.1.STEM.1> & girt \\
<SAMPLE.1.STEM.2> & gir \\
<SAMPLE.1.FORM> & degirim \\
<SAMPLE.2.LEMMA> & kuştin \\
<SAMPLE.2.STEM.1> & kuşt \\
<SAMPLE.2.STEM.2> & kuj \\
<SAMPLE.2.FORM> & dekujim \\ \bottomrule
\end{tabular}
}
\end{table}

\newpage
\section{Prompt for Elicitation Question Generation} \label{appendix_prompt_question}

The following prompt is used to generate question templates that can help the linguist create elicitation questions for speakers who are not familiar with grammatical terminology.

\begin{tcolorbox}[colback=lightgray!10, colframe=gray!50, title= B. Prompt for generating English question templates, fonttitle=\small,fontupper=\small]
\texttt{You are a field linguist working with native speakers to study the morphology of their language. The speakers are not trained in linguistics but understand English. Ask a speaker how they would say the word "XXX" with these specific features: "<\textbf{FEATURE-VALUES SET PLACEHOLDER}>".
Generate only the question for the speaker. "XXX" should be in the question. Do not use linguistic terms. Keep it under 50 words. Do not explain anything to me.}
\end{tcolorbox}

For example, the UniMorph tag set ``\texttt{V;PST}'' is substituted with `\textit{`Part-of-Speech=Verb, Tense=Past}'', which will then replace the corresponding placeholder in the prompt.
\end{document}